\documentclass[11pt,a4paper]{article}
\usepackage[hyperref]{emnlp2020}
\usepackage{times}
\usepackage{latexsym}
\usepackage{microtype}
\usepackage{graphicx}
\usepackage{subfigure}
\usepackage{booktabs} 
\usepackage{amssymb}
\usepackage{amsmath}
\usepackage{bm}
\usepackage{hyperref}

\usepackage{microtype}

\usepackage{microtype}

\aclfinalcopy 
\def\aclpaperid{1315} 

\title{CodeBERT:\\ A Pre-Trained Model for Programming and Natural Languages}

\author{ Zhangyin Feng$^{1}$\thanks{ ~ Work done while this author was an intern at Microsoft Research Asia.}, Daya Guo$^{2}$\footnotemark[1], Duyu Tang$^{3}$, Nan Duan$^{3}$, Xiaocheng Feng$^{1}$ \\
{\bf Ming Gong$^{4}$, Linjun Shou$^{4}$, Bing Qin$^{1}$, Ting Liu$^{1}$, Daxin Jiang$^{4}$, Ming Zhou$^{3}$} \\
$^1$ Research Center for Social Computing and Information Retrieval, Harbin Institute of Technology, China \\
$^2$ The School of Data and Computer Science, Sun Yat-sen University, China\\
$^3$ Microsoft Research Asia, Beijing, China\\
$^4$ Microsoft Search Technology Center Asia, Beijing, China\\
\texttt{ \{zyfeng,xcfeng,qinb,tliu\}@ir.hit.edu.cn}  \\
\texttt{ guody5@mail2.sysu.edu.cn } \\
\texttt{\{dutang,nanduan,migon,lisho,djiang,mingzhou\}@microsoft.com} \\}

\date{}

\begin{document}
\maketitle
\begin{abstract}
We present CodeBERT, a \textit{bimodal} pre-trained model for programming language (PL) and natural language (NL).
CodeBERT learns general-purpose representations that support downstream NL-PL applications such as natural language code search, code documentation generation, etc. We develop CodeBERT with Transformer-based neural architecture, and 
train it with a hybrid objective function
that incorporates 
the pre-training task of 
replaced token detection, which is to detect plausible alternatives sampled from generators.
This enables us to utilize both ``\textit{\textit{bimodal}}'' data of NL-PL pairs and ``\textit{\textit{unimodal}}'' data, where the former 
provides input tokens for model training while the latter helps to learn better generators.
We evaluate CodeBERT on two NL-PL applications by fine-tuning model parameters.
Results show that CodeBERT achieves state-of-the-art performance on 
both natural language code search and 
code documentation generation.
Furthermore, to investigate what type of knowledge is learned in CodeBERT, we construct a dataset for NL-PL probing, and evaluate in a zero-shot setting where parameters of pre-trained models are fixed.
Results show that CodeBERT performs better than previous pre-trained models on NL-PL probing.\footnote{ All the codes and data are available at \url{https://github.com/microsoft/CodeBERT}}
\end{abstract}

\section{Introduction}
\label{section:introduction}
Large pre-trained models such as ELMo \cite{peters2018deep}, GPT \cite{radford2018improving}, BERT \cite{devlin2018bert}, XLNet \cite{yang2019xlnet} and RoBERTa \cite{liu2019roberta} have dramatically improved the state-of-the-art on a variety of natural language processing (NLP) tasks.
These pre-trained models learn effective contextual representations from massive unlabeled text optimized by self-supervised objectives, such as masked language modeling, which  predicts the original masked word from an artificially masked input sequence.
The success of pre-trained models in NLP also drives a surge of multi-modal pre-trained models, such as ViLBERT \cite{lu2019vilbert} for language-image and VideoBERT \cite{sun2019videobert} for language-video, which are learned from \textit{bimodal} data such as language-image pairs with \textit{bimodal} self-supervised objectives.

In this work, we present CodeBERT, a \textit{bimodal} pre-trained model for natural language (NL) and programming language (PL) like Python, Java, JavaScript, etc. 
CodeBERT 
captures the semantic connection between natural language and programming language, and produces general-purpose representations that can broadly support NL-PL understanding tasks (e.g. natural language code search) and generation tasks (e.g. code documentation generation). 
It is \mbox{developed} with the multi-layer Transformer  \cite{vaswani2017attention}, which is adopted in a majority of large pre-trained models.
In order to make use of both \textit{\textit{bimodal}} instances of NL-PL pairs and large amount of available \textit{\textit{unimodal}} codes, we train CodeBERT with a hybrid objective function, including standard masked language modeling \cite{devlin2018bert} and replaced token detection \cite{clark2020electra}, where \textit{\textit{unimodal}} codes help to learn better generators for producing better alternative tokens for the latter objective. 

We train CodeBERT from Github code repositories in 6 programming languages, where \textit{bimodal} datapoints are codes that pair with function-level natural language documentations \cite{husain2019codesearchnet}.
Training is conducted in a setting similar to that of multilingual BERT \cite{pires2019multilingual}, in which case one pre-trained model is learned for 6 programming languages with no explicit markers used to denote the input programming language. 
We evaluate CodeBERT on two downstream NL-PL tasks, including natural language code search and code documentation generation.
Results show that fine-tuning the parameters of CodeBERT achieves state-of-the-art performance on both tasks.
To further investigate what type of knowledge is learned in CodeBERT, we construct a dataset for NL-PL probing, and test CodeBERT in a zero-shot scenario, i.e. without fine-tuning the parameters of CodeBERT. We find that CodeBERT consistently outperforms RoBERTa, a purely natural language-based pre-trained model.
The contributions of this work are as follows:
\begin{itemize}
    \item CodeBERT is the first large NL-PL pre-trained model for multiple programming languages. 
    \item Empirical results show that CodeBERT is effective in both code search and code-to-text generation tasks.
    \item We further created a dataset which is the first one to investigate the probing ability of the code-based pre-trained models.
\end{itemize}

\section{Background}
\subsection{Pre-Trained Models in NLP}
Large pre-trained models \cite{peters2018deep,radford2018improving,devlin2018bert,yang2019xlnet,liu2019roberta,raffel2019exploring} have brought dramatic empirical improvements on almost every NLP task in the past few years. 
Successful approaches train deep neural networks on large-scale plain texts with self-supervised learning objectives.
One of the most representative neural architectures is the Transformer \cite{vaswani2017attention}, which is also the one used in this work.
It contains multiple self-attention layers, and can be conventionally learned with gradient decent in an end-to-end manner as every component is differentiable. 
The terminology ``self-supervised'' means that supervisions used for pre-training are automatically collected from raw data without manual annotation. Dominant learning objectives are language modeling and its variations.
For example, in \mbox{GPT} \cite{radford2018improving}, the learning objective is language modeling, namely predicting the next word $w_k$ given the preceding context words \{$w_1, w_2, ..., w_{k-1}$\}.
As the ultimate goal of pre-training is not to train a good language model, it is desirable to consider both preceding and following contexts to learn better general-purpose contextual representations.
This leads us to the masked language modeling objective used in BERT \cite{devlin2018bert}, which learns to predict the masked words of a randomly masked word sequence given surrounding contexts.
Masked language modeling is also used as one of the two learning objectives for training CodeBERT.

\subsection{Multi-Modal Pre-Trained Models}
The remarkable success of the pre-trained model in NLP has driven the development of multi-modal pre-trained model that learns implicit alignment between inputs of different modalities. 
These models are typically learned from \textit{bimodal} data, such as pairs of language-image or pairs of language-video. 
For example, ViLBERT \cite{lu2019vilbert} learns from image caption data, where the model learns by reconstructing categories of masked image region or masked words given the observed inputs, and meanwhile predicting whether the caption describes the image content or not.
Similarly, VideoBERT \cite{sun2019videobert} learns from language-video data and is trained by video and text masked token prediction.
Our work belongs to this line of research
 as we regard NL and PL as different modalities. 
Our method differs from previous works in that the fuels for model training include not only \textit{bimodal} data of NL-PL pairs, but larger amounts of \textit{unimodal} data such as codes without paired documentations.

A concurrent work  \cite{kanade2019pre} uses masked language modeling and next sentence prediction as the objective to train a BERT model on Python source codes, where a sentence is a logical code line as defined by the Python standard. In terms of the pre-training process, CodeBERT differs from their work in that (1) CodeBERT is trained in a cross-modal style and leverages both bimodal NL-PL data and unimodal PL/NL data, (2) CodeBERT is pre-trained over six programming languages, and (3) CodeBERT is trained with a new learning objective based on replaced token detection.

\section{CodeBERT}
We describe the details about CodeBERT in this section, including the model architecture, the input and output representations, the objectives and data used for training CodeBERT, and how to fine-tune CodeBERT when it is applied to downstream tasks.

\subsection{Model Architecture}
We follow BERT \cite{devlin2018bert} and RoBERTa \cite{liu2019roberta}, and use multi-layer bidirectional Transformer \cite{vaswani2017attention} as the model architecture of CodeBERT.
We will not review  the ubiquitous Transformer architecture in detail. We develop CodeBERT by using exactly the same model architecture as RoBERTa-base.
The total number of model parameters is 125M.

\subsection{Input/Output Representations}
In the pre-training phase, we set the input as the concatenation of two segments with a special separator token, namely $[CLS], w_1, w_2, ..w_n, [SEP], \\ c_1, c_2, ..., c_m, [EOS]$.
One segment is natural language text, and another is code from a certain programming language. 
$[CLS]$ is a special token in front of the two segments, whose final hidden representation is considered as the aggregated sequence representation for classification or ranking.
Following the standard way of processing text in Transformer, we regard a natural language text as a sequence of words, and split it as WordPiece \cite{wu2016google}. 
We regard a piece of code as  a sequence of tokens.

The output of CodeBERT includes (1) contextual vector representation of each token, for both natural language and code, and (2) the representation of $[CLS]$, which works as the aggregated sequence representation.

\subsection{Pre-Training Data}
We train CodeBERT with both \textit{bimodal} data, which refers to parallel data of natural language-code pairs, and \textit{unimodal} data, which stands for codes without paired natural language texts and natural language without paired codes. 

\begin{table}[ht]

\begin{center}
\begin{small}
\begin{sc}
\begin{tabular}{lccc}
\toprule
Training Data & \textit{bimodal} data & \textit{unimodal} Codes   \\
\midrule
Go    & 319,256 & 726,768   \\
Java &  500,754 & 1,569,889  \\
JavaScript    & 143,252& 1,857,835  \\
PHP    & 662,907& 977,821        \\
Python     & 458,219& 1,156,085 \\
Ruby      & 52,905& 164,048 \\
All      & 2,137,293& 6,452,446         \\
\bottomrule
\end{tabular}
\end{sc}
\end{small}
\end{center}
\vskip -0.1in
\caption{ \label{table-codebert-training-data-stat} Statistics of the dataset used for training CodeBERT.}
\end{table}

We use datapoints from Github repositories, where each \textit{bimodal} datapoint is an individual function with paired documentation, and each \textit{unimodal} code is a function without paired documentation.
Specifically, we use a recent large dataset provided by \citet{husain2019codesearchnet}, which includes 2.1M \textit{bimodal} datapoints and 6.4M \textit{unimodal} codes across six programming languages (Python, Java, JavaScript, PHP, Ruby, and Go). Data statistics is shown in Table \ref{table-codebert-training-data-stat}.\footnote{Since we will evaluate on the natural language code search task, we only use the training data of \citet{husain2019codesearchnet} to train CodeBERT with no access to the dev and testing data.}

\begin{figure}[ht]
\begin{center}
\centerline{\includegraphics[width=\columnwidth]{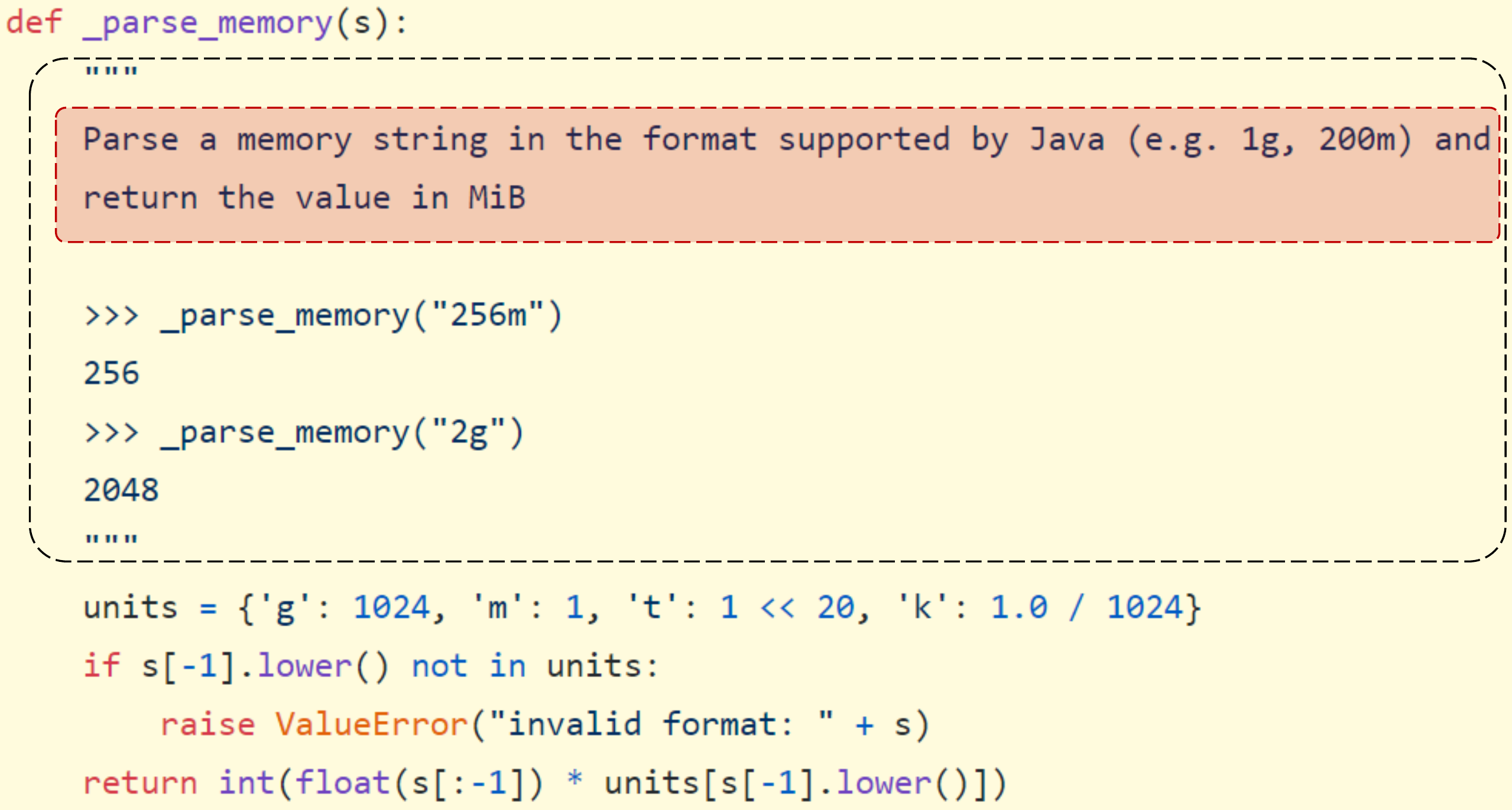}}
\caption{An example of the NL-PL pair, where NL is the first paragraph (filled in red) from the documentation (dashed line in black) of a function.}
\label{figure-example-data}
\end{center}
\vskip -0.2in
\end{figure}

The data comes from publicly available open-source non-fork GitHub repositories and are filtered with a set of constraints and rules. 
For example, (1) each project should be used by at least one other project, (2) each documentation is truncated to the first paragraph, (3) documentations shorter than three tokens are removed, (4) functions shorter than three lines are removed, and (5) function names with substring ``\textit{test}'' are removed.
An example of the data is given in Figure \ref{figure-example-data}
\footnote{The source of the illustrating example comes from \url{https://github.com/apache/spark/blob/618d6bff71073c8c93501ab7392c3cc579730f0b/python/pyspark/rdd.py\#L125-L138}}.

\begin{figure*}
    \centering
    \resizebox{0.94\linewidth}{!}{
    \includegraphics{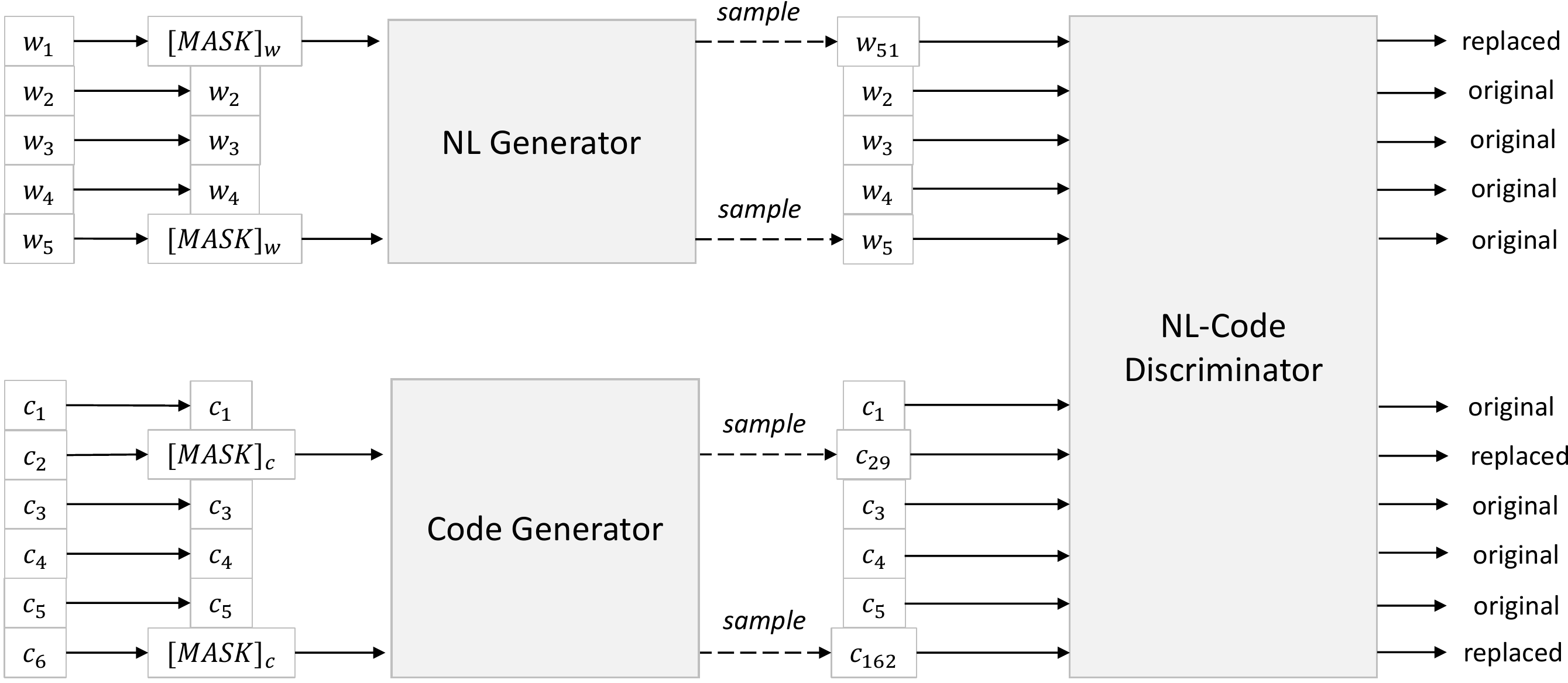}
    }
    \caption{An illustration about the replaced token detection objective. 
    Both NL and code generators are language models, which generate plausible tokens for masked positions based on surrounding contexts.  
    NL-Code discriminator is the targeted pre-trained model, which is trained via detecting plausible alternatives tokens sampled from NL and PL generators. NL-Code discriminator is used for producing general-purpose representations in the fine-tuning step. 
    Both NL and code generators are thrown out in the fine-tuning step. }
    \label{fig:my_label}
\end{figure*}

\subsection{Pre-Training CodeBERT}
We describe the two objectives used for training CodeBERT here. The first objective is masked language modeling (MLM), which has proven effective in literature \cite{devlin2018bert,liu2019roberta,sun2019videobert}. 
We apply masked language modeling on \textit{bimodal} data of NL-PL pairs.
The second objective is replaced token detection (RTD), which further uses a large amount of \textit{unimodal} data, such as codes without paired natural language texts. Detailed hyper-parameters for model pre-training are given in Appendix B.1.

\paragraph{Objective \#1: Masked Language Modeling (MLM)}
Given a datapoint of NL-PL pair ($\bm{x}=\{\bm{w}$, $\bm{c}$\}) as input, where $\bm{w}$ is a sequence of NL words and  $\bm{c}$ is a sequence of PL tokens,
we first select a random set of positions for both NL and PL to mask out (i.e. $\bm{m^w}$ and $\bm{m^c}$, respectively), and then replace the selected positions with a special $[MASK]$ token. Following \citet{devlin2018bert}, 15\% of the tokens from $\bm{x}$ are masked out.
\begin{align}
    m_i^w \sim& \text{ unif}\{1,|\bm{w}|\}\text{ for }i = 1\text{ to }|\bm{w}|\\
    m_i^c \sim& \text{ unif}\{1,|\bm{c}|\}\text{ for }i = 1\text{ to }|\bm{c}|\\
    \bm{w}^{\text{masked}} = &\text{ REPLACE}(\bm{w}, \bm{m^w}, [MASK])\\
    \bm{c}^{\text{masked}} = &\text{ REPLACE}(\bm{c}, \bm{m^c}, [MASK])\\
    \bm{x} = &\ \bm{w} + \bm{c}
\end{align}
The MLM objective is to predict the original tokens which are masked out, formulated as follows, where $p^{D_1}$ is the discriminator which predicts a token from a large vocabulary.
\begin{equation}
\centering
    \mathcal{L}_{\text{MLM}}(\theta)\! = \!\!\!
     \sum_{i \in \bm{m^w} \cup \bm{m^c}} \!\!\!
    -\text{log } p^{D_1}(x_i |\! \bm{w}^{\text{masked}},\! \bm{c}^{\text{masked}})
\end{equation}

\paragraph{Objective \#2: Replaced Token Detection (RTD)}
In the MLM objective, only \textit{bimodal} data (i.e. datapoints of NL-PL pairs) is used for training. 
Here we present the objective of replaced token detection.
The RTD objective \cite{clark2020electra} is originally developed for efficiently learning pre-trained model for natural language. We adapt it in our scenario, with the advantage of using both \textit{bimodal} and \textit{unimodal} data for training. 
Specifically, there are two data generators here, an NL generator $p^{G_w}$ and a PL generator $p^{G_c}$, both for generating plausible alternatives for the set of randomly masked positions. 
\begin{align}
    \hat{w_i} &\sim p^{G_w}(w_i|\bm{w}^{\text{masked}}) \text{ for }i \in \bm{m^w}\\
    \hat{c_i} &\sim p^{G_c}(c_i|\bm{c}^{\text{masked}}) \text{ for }i \in \bm{m^c}
\end{align}
\begin{align}
    \bm{w}^{\text{corrupt}} = &\text{ REPLACE}(\bm{w}, \bm{m^w}, \hat{\bm{w}}) \\
    \bm{c}^{\text{corrupt}} = &\text{ REPLACE}(\bm{c}, \bm{m^c}, \hat{\bm{c}})\\
     \bm{x}^{\text{corrupt}} = &\ \bm{w}^{\text{corrupt}} +  \bm{c}^{\text{corrupt}}
\end{align}
The discriminator is trained to determine whether a word is the original one or not, which is a binary classification problem.
It is worth noting that the RTD objective is applied to every position in the input, and it differs from GAN (generative adversarial network) in that if a generator happens to produce the correct token, the label of that token is ``real'' instead of ``fake'' \cite{clark2020electra}.
The loss function of RTD with regard to the discriminator parameterized by $\theta$ is given below, where $\delta(i)$ is an indicator function and $p^{D_2}$ is the discriminator that predicts the probability of the $i$-th word being original. 
\begin{align}
    \mathcal{L}_{\text{RTD}}(\theta) = &\sum_{i = 1}^{|\bm{w}|+|\bm{c}|} \bigg( \delta(i) \text{log }p^{D_2}(\bm{x}^{\text{corrupt}}, i) + \nonumber\\ &\Big(1-\delta(i)\Big)\Big(1- \text{log }p^{D_2}(\bm{x}^{\text{corrupt}}, i)\Big) \bigg)
\end{align}
\begin{equation}
  \delta(i)=\begin{cases}
    1, & \text{if $x_i^{\text{corrupt}}=x_i$}.\\
    0, & \text{otherwise}.
  \end{cases}
\end{equation}

There are many different ways to implement the generators. 
In this work, we implement two efficient n-gram language models \cite{jurafsky2000speech} with bidirectional contexts, one for NL and one for PL, and learn them from corresponding unimodel datapoints, respectively. 
The approach is easily generalized to learn \textit{bimodal} generators or use more complicated generators like Transformer-based neural architecture learned in a joint manner. We leave these to future work. 
The PL training data is the \textit{unimodal} codes as shown in Table \ref{table-codebert-training-data-stat}, and the NL training data comes from the documentations from \textit{bimodal} data. One could easily extend these two training datasets to larger amount.
The final loss function are given below.
\begin{align}
    \min_{\theta} 
    \mathcal{L}_{\text{MLM}}(\theta) + \mathcal{L}_{\text{RTD}}(\theta) 
\end{align}

\subsection{Fine-Tuning CodeBERT}

We have different settings to use CodeBERT in downstream NL-PL tasks. For example, in natural language code search, we feed the input as the same way as the pre-training phase and use the representation of $[CLS]$ to measure the semantic relevance between code and natural language query, while in code-to-text generation, we use an encoder-decoder framework and initialize the encoder of a generative model with CodeBERT. 
Details are given in the experiment section.

\section{Experiment}

We present empirical results in this section to verify the effectiveness of CodeBERT. 
We first describe the use of CodeBERT in natural language code search (\S \ref{section:experiment-code-search}), in a way that model parameters of CodeBERT are fine-tuned. 
After that, we present the NL-PL probing task (\S \ref{section:experiment-nl-pl-probing}), and evaluate CodeBERT in a zero-shot setting where the parameters of CodeBERT are fixed.
Finally, we evaluate CodeBERT on a generation problem, i.e. code documentation generation (\S \ref{section:experiment-code-documentation-generation}), and further evaluate on a programming language which is never seen in the training phase (\S \ref{section:generalization-to-csharp}). 

\begin{table*}

\begin{center}
\begin{small}
\begin{sc}
\begin{tabular}{p{4.8cm}ccccccc}
\toprule
model & ruby & javascript & go & python & java & php & Ma-Avg\\
\midrule
NBow & 0.4285 & 0.4607 & 0.6409 & 0.5809 & 0.5140 & 0.4835 & 0.5181\\
CNN & 0.2450 & 0.3523 & 0.6274 & 0.5708 & 0.5270 & 0.5294 & 0.4753\\
BiRNN & 0.0835 & 0.1530 & 0.4524 & 0.3213 & 0.2865 & 0.2512 & 0.2580\\
selfAtt & 0.3651 & 0.4506 & 0.6809 & 0.6922 & 0.5866 & 0.6011 & 0.5628\\
\hline
RoBerta & 0.6245 & 0.6060 & 0.8204 & 0.8087 & 0.6659 & 0.6576 & 0.6972\\
PT w/ Code Only (init=s) & 0.5712	& 0.5557 & 0.7929 &	0.7855 & 0.6567	& 0.6172 & 0.6632 \\
PT w/ Code Only (init=R) & 0.6612 & 0.6402 & 0.8191 & 0.8438 & 0.7213 & 0.6706 & 0.7260 \\
\hline
CodeBERT (MLM, init=s) & 0.5695 & 0.6029 & 0.8304 & 0.8261 & 0.7142 & 0.6556 & 0.6998\\
CodeBERT (MLM, init=R)  & 0.6898 & 0.6997 & 0.8383 & 0.8647 & 0.7476 & 0.6893 & 0.7549\\
CodeBERT (RTD, init=R) & 0.6414 & 0.6512 & 0.8285 & 0.8263 & 0.7150 & 0.6774 & 0.7233\\
CodeBERT~(MLM+RTD, init=R) & \textbf{0.6926} & \textbf{0.7059} & \textbf{0.8400} & \textbf{0.8685} & \textbf{0.7484} & \textbf{0.7062} & \textbf{0.7603}\\
\bottomrule
\end{tabular}
\end{sc}
\end{small}
\end{center}
\caption{ \label{table-codesearchnet-results} Results on natural language code retrieval. Baselines include four  joint embeddings (first group) of NL and PL, RoBERTa, and RoBERTa which is continuously trained with masked language modeling on codes only (second group). \textsc{PT} stands for pre-training. We train CodeBERT (third group) with different settings, including using different initialization (from scratch ({\textsc{init=s}}) or initialized with the parameters of RoBERTa ({\textsc{init=R}})) and using different learning objectives (MLM, RTD, or the combination of both).}
\end{table*}
\subsection{Natural Language Code Search}\label{section:experiment-code-search}
Given a natural language as the input, the objective of code search is to find the most semantically related code from a collection of codes.
We conduct experiments on the CodeSearchNet corpus \cite{husain2019codesearchnet}
\footnote{More details about the dataset are given in Appendix A.}.
We follow the official evaluation metric to calculate the Mean Reciprocal Rank (MRR) for each pair of test data ($\bm{c}$, $\bm{w}$) over a fixed set of 999 distractor codes. We further calculate the macro-average MRR for all languages as an overall evaluation metric.
It is helpful to note that this metric differs from the \textsc{avg} metric in the original paper, where the answer is retrieved from candidates from all six languages.
We fine-tune a language-specific model for each programming language\footnote{We have fine-tuned a multi-lingual model for six programming languages, but find that it performs worse that fine-tuning a  language-specific model for each programming language.}.
We train each model with a binary classification loss function, where a $softmax$ layer is connected to the representation of $[CLS]$. 
Both training and validation datasets are created in a way that positive and negative samples are balanced. Negative samples consist of balanced number of instances with randomly replaced NL (i.e. ($\bm{c}$, $\bm{\hat{w}}$))  and PL (i.e. ($\bm{\hat{c}}$, $\bm{w}$)). Detailed hyper-parameters
for model fine-tuning are given in Appendix B.2.

\paragraph{Model Comparisons}

Table \ref{table-codesearchnet-results} shows the results of different approaches on the CodeSearchNet corpus.
The first four rows are reported by \citet{husain2019codesearchnet}, which are joint embeddings of NL and PL \cite{gu2018deep,mitra2018introduction}.
\textbf{\textsc{NBoW}} represents neural bag-of-words. \textbf{\textsc{CNN}}, \textbf{\textsc{BiRNN}} and \textbf{\textsc{SelfATT}}  stand for 1D convolultional  neural network \cite{kim2014convolutional}, bidirectional GRU-based recurrent neural network \cite{cho2014learning}, and multi-head attention \cite{vaswani2017attention}, respectively. 

We report the remaining numbers in Table \ref{table-codesearchnet-results}. We train all these pre-trained models by regarding codes as a sequence of tokens. 
We also continuously train RoBERTa only on codes from CodeSearchNet with masked language modeling.
Results show that CodeBERT consistently performs better than RoBERTa and the model pre-trained with code only.
CodeBERT (MLM) learned from scratch performs better than RoBERTa. 
Unsurprisingly, initializing CodeBERT with RoBERTa improves the performance
\footnote{We further give a learning curve of different pre-trained models in the fine-tuning process in Appendix C.}. 

\subsection{NL-PL Probing}\label{section:experiment-nl-pl-probing}
In the previous subsection, we show the empirical effectiveness of CodeBERT in a setting that the parameters of CodeBERT are fine-tuned in downstream tasks. 
In this subsection, we further investigate what type of knowledge is learned in CodeBERT without modifying the parameters.

\paragraph{Task Formulation and Data Construction}
Following the probing experiments in NLP \cite{petroni2019language,talmor2019olmpics}, we study NL-PL probing here.
Since there is no existing work towards this goal, we formulate the problem of NL-PL probing and create the dataset by ourselves.
Given an NL-PL pair ($\bm{c}$, $\bm{w}$), the goal of NL-PL probing is to test model's ability to correctly predict/recover the masked token of interest (either a code token $c_i$ or word token $w_j$) among distractors. 
There are two major types of distractors: one is the whole target vocabulary used for the masked language modeling objective \cite{petroni2019language}, and another one has fewer candidates which are filter or curated based on experts' understanding about the ability to be tested \cite{talmor2019olmpics}. 
We follow the second direction and formulate NL-PL probing as a  multi-choice question answering task, where the question is cloze-style in which a certain token is replaced by $[MASK]$ and distractor candidate answers are curated based on our expertise. 

Specifically, we evaluate on the NL side and PL side, respectively. 
To ease the effort of data collection, we collect data automatically from NL-PL pairs in both validation and testing sets of CodeSearchNet, both of which are unseen in the pre-training phase. 
To evaluate on the NL side, we select NL-PL pairs whose NL documentations include one of the six keywords (\textit{max}, \textit{maximize}, \textit{min}, \textit{minimize}, \textit{less}, \textit{greater}), and group them to four candidates by merging first two keywords and the middle two keywords. The task is to ask pre-trained models to select the correct one instead of three other distractors. That is to say, the input in this setting includes the complete code and a masked NL documentation. The goal is to select the correct answer from four candidates.
For the PL side, we select codes containing keywords \textit{max} and \textit{min}, and formulate the task as a two-choice answer selection problem.
Here, the input includes complete NL documentation and a masked PL code, and the goal is to select the correct answer from two candidates. Since code completion is an important scenario, we would like to test model's ability in predicting the correct token merely based on preceding PL contexts. Therefore, we add an additional setting for PL side, where the input includes the complete NL documentation and preceding PL codes. 
Data statistics is given in the top two rows in Table~\ref{table-probing-results}.
\begin{table*}
	\begin{center}
		\begin{small}
			\begin{sc}
				\begin{tabular}{lccccccc}
					\toprule
					& ruby & javascript & go & python & java & php & all\\
					\midrule
					\multicolumn{8}{l}{\textbf{number of datapoints for probing}} \\
					PL (2 choices) & 38 & 272 & 152 & 1,264 & 482 & 407 & 2,615\\
					NL (4 choices) & 20 & 65 & 159 & 216 & 323 & 73 & 856\\
					\bottomrule
					\multicolumn{8}{l}{\textbf{PL probing}}\\
					Roberta  & 73.68 & 63.97 & 72.37 & 59.18 & 59.96 & 69.78 & 62.45\\
					Pre-Train w/ Code Only & 71.05 & 77.94 & 89.47 & 70.41 & 70.12 & 82.31 & 74.11\\
					CodeBERT (MLM) & \textbf{86.84} & \textbf{86.40} & \textbf{90.79} & \textbf{82.20} & \textbf{90.46} & \textbf{88.21} & \textbf{85.66}\\
					\bottomrule
					\multicolumn{2}{l}{\textbf{PL probing with preceding context only}}\\
					Roberta & \textbf{73.68} & \textbf{53.31} & 51.32 & 55.14 & 42.32 & 52.58 & 52.24\\
					Pre-Train w/ Code Only & 63.16 & 48.53 & \textbf{61.84} & 56.25 & \textbf{58.51} & 58.97 & 56.71\\
					CodeBERT (MLM) & 65.79 & 50.74 & 59.21 & \textbf{62.03} & 54.98 & \textbf{59.95} & \textbf{59.12}\\
					\bottomrule
					\multicolumn{8}{l}{\textbf{NL probing}}\\
					Roberta & 50.00\ & 72.31 & 54.72 & 61.57 & 61.61 & 65.75 & 61.21\\
					Pre-Train w/ Code Only & 55.00 & 67.69 & 60.38 & 68.06 & 65.02 & 68.49 & 65.19\\
					CodeBERT (MLM) & \textbf{65.00} & \textbf{89.23} & \textbf{66.67} & \textbf{76.85} & \textbf{73.37} & \textbf{79.45} & \textbf{74.53}\\
					\bottomrule
				\end{tabular}
			\end{sc}
		\end{small}
	\end{center}
	\vskip -0.1in
	\caption{ \label{table-probing-results} Statistics of the data for NL-PL probing and the performance of different pre-trained models. Accuracies (\%) are reported. Best results in each group are in bold.}
\end{table*}

\paragraph{Model Comparisons} Results are given in Table \ref{table-probing-results}. We report 
accuracy, namely the number of correctly predicted instances over the number of all instances, for each programming language. 
Since datasets in different programming languages are extremely unbalanced, we report the accumulated metric 
with the same way.
We use CodeBERT (MLM) here because its output layer naturally fits for probing.
Results show that CodeBERT performs better than baselines on almost all languages on both NL and PL probing.
The numbers with only preceding contexts are lower than that with bidirectional contexts, which suggests that code completion is challenging. We leave it as a future work.

\begin{figure}[t]
\begin{center}
\centerline{\includegraphics[width=\columnwidth]{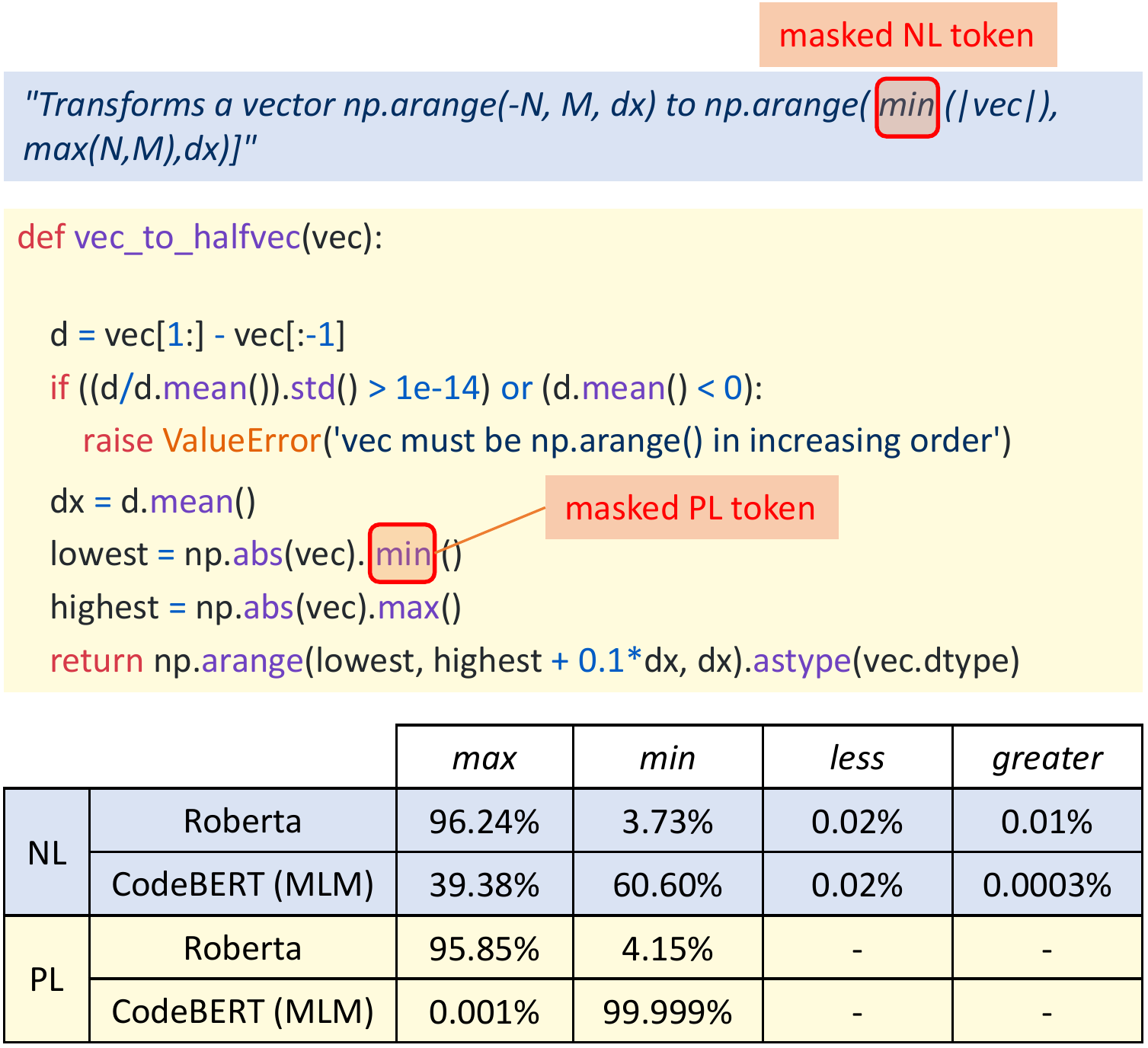}}
\caption{Case study on python language. Masked tokens in NL (in blue) and PL (in yellow) are separately applied. Predicted probabilities of RoBERTa and CodeBERT are given. }
\label{figure-nl-pl-probing-examplep}
\end{center}
\vskip -0.35in
\end{figure}

We further give a case study on PL-NL probing.
We mask NL token and PL token separately, and report the predicted probabilities of RoBERTa and CodeBERT. 
Figure~\ref{figure-nl-pl-probing-examplep} illustrates the example of a python code\footnote{The example comes from \url{https://github.com/peri-source/peri/blob/61beed5deaaf978ab31ed716e8470d86ba639867/peri/comp/psfcalc.py\#L994-L1002}}.
We can see that RoBERTa fails in both cases, whereas CodeBERT makes the correct prediction in both NL and PL settings. 

\begin{table*}

\begin{center}
\begin{small}
\begin{sc}
\begin{tabular}{lccccccc}
\toprule
model & ruby & javascript & go & python & java & php & overall\\
\midrule
seq2seq & 9.64 & 10.21 & 13.98 & 15.93 & 15.09 & 21.08 & 14.32\\
Transformer & 11.18 & 11.59 & 16.38 & 15.81 & 16.26 & 22.12 & 15.56\\
RoBERTa & 11.17 & 11.90 & 17.72 & 18.14 & 16.47 & 24.02 & 16.57\\
pre-train w/ code only & 11.91 & 13.99 & 17.78 & 18.58 & 17.50 & 24.34 & 17.35\\
CodeBERT (rtd) & 11.42 & 13.27 & 17.53 & 18.29 & 17.35&24.10 & 17.00\\
CodeBERT (mlm) & 11.57 & 14.41 & 17.78 & 18.77 & 17.38 & 24.85 & 17.46\\
CodeBERT (rtd+mlm) & {\bf 12.16} & {\bf 14.90} & \bf{18.07} & {\bf 19.06} & {\bf 17.65} & {\bf 25.16} & {\bf 17.83}\\
\bottomrule
\end{tabular}
\end{sc}
\end{small}
\end{center}
\vskip -0.15in
\caption{\label{table-code2nl-reults-codesearchnet} Results on Code-to-Documentation generation, evaluated with smoothed BLEU-4 score.}
\end{table*}

\subsection{Code Documentation Generation}\label{section:experiment-code-documentation-generation}

Although the pre-training objective of CodeBERT does not include generation-based objectives \cite{lewis2019bart}, we would like to investigate to what extent does CodeBERT perform on generation tasks.
Specifically, we study code-to-NL generation, and report results for the documentation generation task on CodeSearchNet Corpus in six programming languages. 
Since the generated documentations are short and higher order n-grams may not overlap, we remedy this problem by using smoothed BLEU score \cite{lin2004orange}. 

\paragraph{Model Comparisons}
We compare our model with several baselines, including a RNN-based model with attention mechanism \cite{sutskever2014sequence}, the Transformer \cite{vaswani2017attention}, RoBERTa and the model pre-trained on code only. To demonstrate the effectiveness of CodeBERT on code-to-NL generation tasks, we  adopt various pre-trained models as encoders and keep the hyperparameters consistent. Detailed hyper-parameters are given in Appendix B.3.

Table \ref{table-code2nl-reults-codesearchnet} shows the results with different models for the code-to-documentation generation task. As we can see, models pre-trained on programming language outperform RoBERTa, which illustrates that pre-trainning models on programming language could improve code-to-NL generation. Besides, results in the Table \ref{table-code2nl-reults-codesearchnet} show that CodeBERT pre-trained with RTD and MLM objectives brings a gain of 1.3 BLEU score over RoBERTa overall and achieve the state-of-the-art performance\footnote{We further give some output examples in Appendix E.}.


\subsection{Generalization to Programming Languages NOT in
	Pre-training}\label{section:generalization-to-csharp}

We would like to evaluate CodeBERT on the programming language which is never seen in the pre-training step.
To this end, we study the task of generating a natural language summary of a C\# code snippet. 
We conduct experiments on the dataset of CodeNN \cite{iyer2016summarizing}\footnote{\url{https://github.com/sriniiyer/codenn}}, which consists of 66,015 pairs of questions and answers automatically collected from StackOverflow.
This dataset is challenging since the scale of dataset is orders of magnitude smaller than CodeSearchNet Corpus.
We evaluate models using smoothed BLEU-4 score and use the same evaluation scripts as \citet{iyer2016summarizing}. 

\begin{table}[ht]
	\begin{center}
		\begin{small}
			\begin{sc}
				\begin{tabular}{lc}
					\toprule
					Model & BLEU   \\
					\midrule
					MOSES \cite{koehn2007moses}& 11.57\\
					IR &13.66\\
					SUM-NN \cite{rush2015neural}& 19.31\\
					2-layer BiLSTM& 19.78\\
					Transformer \cite{vaswani2017attention}& 19.68\\
					TreeLSTM \cite{tai2015improved} &20.11\\
					CodeNN \cite{iyer2016summarizing}& 20.53\\
					code2seq \cite{alon2018code2seq}& {\bf 23.04}\\
					\hline
					RoBERTa & 19.81\\
					pre-train w/ code only & 20.65\\ 
					CodeBERT (RTD) & 22.14\\
					CodeBERT (MLM) & 22.32\\
					
					CodeBERT (MLM+RTD) & {\bf 22.36}\\
					\bottomrule
				\end{tabular}
			\end{sc}
		\end{small}
	\end{center}
	\vskip -0.1in
	\caption{\label{table-code2nl-c} Code-to-NL generation on C\# language.}
\end{table}

\paragraph{Model Comparisons}
Table \ref{table-code2nl-c} shows that our model with MLM and RTD pre-training objectives achieves 22.36 BLEU score and improves by 2.55 points over RoBERTa, which illustrates CodeBERT could generalize better to other programming language which is never seen in the pre-training step. However, our model achieve slightly lower results than code2seq \cite{alon2018code2seq}. The main reason could be that code2seq makes use of compositional paths in its abstract syntax tree (AST) while CodeBERT only takes original code as the input.
We have trained a version of CodeBERT by traversing the tree structure of AST following a certain order, but applying that model does not bring improvements on generation tasks. This shows a potential direction to improve CodeBERT by incorporating AST.

\section{Conclusion}
In this paper, we present CodeBERT, which to the best of our knowledge is the first large \textit{bimodal} pre-trained model for natural language and programming language. 
We train CodeBERT on both \textit{bimodal} and \textit{unimodal} data, and show that fine-tuning CodeBERT achieves state-of-the-art performance on downstream tasks including natural language code search and code-to-documentation generation.
To further investigate the knowledge embodied in pre-trained models, we formulate the task of NL-PL probing and create a dataset for probing. We regard the probing task as a cloze-style answer selection problem, and curate distractors for both NL and PL parts. Results show that, with model \mbox{parameters} fixed, CodeBERT performs better than RoBERTa and a continuously trained model using codes only.

There are many potential directions for further research on this field.
First, one could learn better generators with \textit{bimodal} evidence or more complicated neural architecture to improve the replaced token detection objective.
Second, the loss functions of CodeBERT mainly target on NL-PL understanding tasks. Although CodeBERT achieves strong BLEU scores on code-to-documentation generation, the CodeBERT itself could be further improved by generation-related learning objectives. How to successfully incorporate AST into the pre-training step is also an attractive direction.
Third, we plan to apply CodeBERT to more NL-PL related tasks, and extend it to more programming languages. Flexible and powerful domain/language adaptation methods are necessary to generalize well.

\section*{Acknowledgments}
Xiaocheng Feng is the corresponding author of this work. We thank the anonymous reviewers for their insightful comments.
Zhangyin Feng, Xiaocheng Feng, Bing Qin and Ting Liu are supported by the National Key R\&D Program of China via grant 2018YFB1005103 and National Natural Science Foundation of China (NSFC) via grant 61632011 and 61772156.

\bibliography{emnlp2020}

\begin{thebibliography}{29}
\expandafter\ifx\csname natexlab\endcsname\relax\def\natexlab#1{#1}\fi

\bibitem[{Alon et~al.(2019)Alon, Brody, Levy, and Yahav}]{alon2018code2seq}
Uri Alon, Shaked Brody, Omer Levy, and Eran Yahav. 2019.
\newblock code2seq: Generating sequences from structured representations of
  code.
\newblock \emph{International Conferenceon Learning Representations}.

\bibitem[{Cho et~al.(2014)Cho, Van~Merri{\"e}nboer, Gulcehre, Bahdanau,
  Bougares, Schwenk, and Bengio}]{cho2014learning}
Kyunghyun Cho, Bart Van~Merri{\"e}nboer, Caglar Gulcehre, Dzmitry Bahdanau,
  Fethi Bougares, Holger Schwenk, and Yoshua Bengio. 2014.
\newblock Learning phrase representations using rnn encoder-decoder for
  statistical machine translation.
\newblock \emph{arXiv preprint arXiv:1406.1078}.

\bibitem[{Clark et~al.(2020)Clark, Luong, Le, and Manning}]{clark2020electra}
Kevin Clark, Minh-Thang Luong, Quoc~V. Le, and Christopher~D. Manning. 2020.
\newblock {\{}ELECTRA{\}}: Pre-training text encoders as discriminators rather
  than generators.
\newblock In \emph{International Conference on Learning Representations}.

\bibitem[{Devlin et~al.(2018)Devlin, Chang, Lee, and
  Toutanova}]{devlin2018bert}
Jacob Devlin, Ming-Wei Chang, Kenton Lee, and Kristina Toutanova. 2018.
\newblock Bert: Pre-training of deep bidirectional transformers for language
  understanding.
\newblock \emph{arXiv preprint arXiv:1810.04805}.

\bibitem[{Gu et~al.(2018)Gu, Zhang, and Kim}]{gu2018deep}
Xiaodong Gu, Hongyu Zhang, and Sunghun Kim. 2018.
\newblock Deep code search.
\newblock In \emph{2018 IEEE/ACM 40th International Conference on Software
  Engineering (ICSE)}, pages 933--944. IEEE.

\bibitem[{Husain et~al.(2019)Husain, Wu, Gazit, Allamanis, and
  Brockschmidt}]{husain2019codesearchnet}
Hamel Husain, Ho-Hsiang Wu, Tiferet Gazit, Miltiadis Allamanis, and Marc
  Brockschmidt. 2019.
\newblock Codesearchnet challenge: Evaluating the state of semantic code
  search.
\newblock \emph{arXiv preprint arXiv:1909.09436}.

\bibitem[{Iyer et~al.(2016)Iyer, Konstas, Cheung, and
  Zettlemoyer}]{iyer2016summarizing}
Srinivasan Iyer, Ioannis Konstas, Alvin Cheung, and Luke Zettlemoyer. 2016.
\newblock Summarizing source code using a neural attention model.
\newblock In \emph{Proceedings of the 54th Annual Meeting of the Association
  for Computational Linguistics (Volume 1: Long Papers)}, pages 2073--2083.

\bibitem[{Jurafsky(2000)}]{jurafsky2000speech}
Dan Jurafsky. 2000.
\newblock \emph{Speech \& language processing}.
\newblock Pearson Education India.

\bibitem[{Kanade et~al.(2019)Kanade, Maniatis, Balakrishnan, and
  Shi}]{kanade2019pre}
Aditya Kanade, Petros Maniatis, Gogul Balakrishnan, and Kensen Shi. 2019.
\newblock Pre-trained contextual embedding of source code.
\newblock \emph{arXiv preprint arXiv:2001.00059}.

\bibitem[{Kim(2014)}]{kim2014convolutional}
Yoon Kim. 2014.
\newblock Convolutional neural networks for sentence classification.
\newblock \emph{arXiv preprint arXiv:1408.5882}.

\bibitem[{Koehn et~al.(2007)Koehn, Hoang, Birch, Callison-Burch, Federico,
  Bertoldi, Cowan, Shen, Moran, Zens et~al.}]{koehn2007moses}
Philipp Koehn, Hieu Hoang, Alexandra Birch, Chris Callison-Burch, Marcello
  Federico, Nicola Bertoldi, Brooke Cowan, Wade Shen, Christine Moran, Richard
  Zens, et~al. 2007.
\newblock Moses: Open source toolkit for statistical machine translation.
\newblock In \emph{Proceedings of the 45th annual meeting of the association
  for computational linguistics companion volume proceedings of the demo and
  poster sessions}, pages 177--180.

\bibitem[{Lewis et~al.(2019)Lewis, Liu, Goyal, Ghazvininejad, Mohamed, Levy,
  Stoyanov, and Zettlemoyer}]{lewis2019bart}
Mike Lewis, Yinhan Liu, Naman Goyal, Marjan Ghazvininejad, Abdelrahman Mohamed,
  Omer Levy, Ves Stoyanov, and Luke Zettlemoyer. 2019.
\newblock Bart: Denoising sequence-to-sequence pre-training for natural
  language generation, translation, and comprehension.
\newblock \emph{arXiv preprint arXiv:1910.13461}.

\bibitem[{Lin and Och(2004)}]{lin2004orange}
Chin-Yew Lin and Franz~Josef Och. 2004.
\newblock Orange: a method for evaluating automatic evaluation metrics for
  machine translation.
\newblock In \emph{Proceedings of the 20th international conference on
  Computational Linguistics}, page 501. Association for Computational
  Linguistics.

\bibitem[{Liu et~al.(2019)Liu, Ott, Goyal, Du, Joshi, Chen, Levy, Lewis,
  Zettlemoyer, and Stoyanov}]{liu2019roberta}
Yinhan Liu, Myle Ott, Naman Goyal, Jingfei Du, Mandar Joshi, Danqi Chen, Omer
  Levy, Mike Lewis, Luke Zettlemoyer, and Veselin Stoyanov. 2019.
\newblock Roberta: A robustly optimized bert pretraining approach.
\newblock \emph{arXiv preprint arXiv:1907.11692}.

\bibitem[{Lu et~al.(2019)Lu, Batra, Parikh, and Lee}]{lu2019vilbert}
Jiasen Lu, Dhruv Batra, Devi Parikh, and Stefan Lee. 2019.
\newblock Vilbert: Pretraining task-agnostic visiolinguistic representations
  for vision-and-language tasks.
\newblock In \emph{Advances in Neural Information Processing Systems}, pages
  13--23.

\bibitem[{Mitra et~al.(2018)Mitra, Craswell et~al.}]{mitra2018introduction}
Bhaskar Mitra, Nick Craswell, et~al. 2018.
\newblock An introduction to neural information retrieval.
\newblock \emph{Foundations and Trends{\textregistered} in Information
  Retrieval}, 13(1):1--126.

\bibitem[{Peters et~al.(2018)Peters, Neumann, Iyyer, Gardner, Clark, Lee, and
  Zettlemoyer}]{peters2018deep}
Matthew~E Peters, Mark Neumann, Mohit Iyyer, Matt Gardner, Christopher Clark,
  Kenton Lee, and Luke Zettlemoyer. 2018.
\newblock Deep contextualized word representations.
\newblock \emph{arXiv preprint arXiv:1802.05365}.

\bibitem[{Petroni et~al.(2019)Petroni, Rockt{\"a}schel, Lewis, Bakhtin, Wu,
  Miller, and Riedel}]{petroni2019language}
Fabio Petroni, Tim Rockt{\"a}schel, Patrick Lewis, Anton Bakhtin, Yuxiang Wu,
  Alexander~H Miller, and Sebastian Riedel. 2019.
\newblock Language models as knowledge bases?
\newblock \emph{arXiv preprint arXiv:1909.01066}.

\bibitem[{Pires et~al.(2019)Pires, Schlinger, and
  Garrette}]{pires2019multilingual}
Telmo Pires, Eva Schlinger, and Dan Garrette. 2019.
\newblock How multilingual is multilingual bert?
\newblock \emph{arXiv preprint arXiv:1906.01502}.

\bibitem[{Radford et~al.(2018)Radford, Narasimhan, Salimans, and
  Sutskever}]{radford2018improving}
Alec Radford, Karthik Narasimhan, Tim Salimans, and Ilya Sutskever. 2018.
\newblock Improving language understanding by generative pre-training.
\newblock \emph{URL https://s3-us-west-2. amazonaws.
  com/openai-assets/researchcovers/languageunsupervised/language understanding
  paper. pdf}.

\bibitem[{Raffel et~al.(2019)Raffel, Shazeer, Roberts, Lee, Narang, Matena,
  Zhou, Li, and Liu}]{raffel2019exploring}
Colin Raffel, Noam Shazeer, Adam Roberts, Katherine Lee, Sharan Narang, Michael
  Matena, Yanqi Zhou, Wei Li, and Peter~J Liu. 2019.
\newblock Exploring the limits of transfer learning with a unified text-to-text
  transformer.
\newblock \emph{arXiv preprint arXiv:1910.10683}.

\bibitem[{Rush et~al.(2015)Rush, Chopra, and Weston}]{rush2015neural}
Alexander~M Rush, Sumit Chopra, and Jason Weston. 2015.
\newblock A neural attention model for abstractive sentence summarization.
\newblock \emph{arXiv preprint arXiv:1509.00685}.

\bibitem[{Sun et~al.(2019)Sun, Myers, Vondrick, Murphy, and
  Schmid}]{sun2019videobert}
Chen Sun, Austin Myers, Carl Vondrick, Kevin Murphy, and Cordelia Schmid. 2019.
\newblock Videobert: A joint model for video and language representation
  learning.
\newblock \emph{arXiv preprint arXiv:1904.01766}.

\bibitem[{Sutskever et~al.(2014)Sutskever, Vinyals, and
  Le}]{sutskever2014sequence}
Ilya Sutskever, Oriol Vinyals, and Quoc~V Le. 2014.
\newblock Sequence to sequence learning with neural networks.
\newblock In \emph{Advances in neural information processing systems}, pages
  3104--3112.

\bibitem[{Tai et~al.(2015)Tai, Socher, and Manning}]{tai2015improved}
Kai~Sheng Tai, Richard Socher, and Christopher~D Manning. 2015.
\newblock Improved semantic representations from tree-structured long
  short-term memory networks.
\newblock \emph{arXiv preprint arXiv:1503.00075}.

\bibitem[{Talmor et~al.(2019)Talmor, Elazar, Goldberg, and
  Berant}]{talmor2019olmpics}
Alon Talmor, Yanai Elazar, Yoav Goldberg, and Jonathan Berant. 2019.
\newblock olmpics--on what language model pre-training captures.
\newblock \emph{arXiv preprint arXiv:1912.13283}.

\bibitem[{Vaswani et~al.(2017)Vaswani, Shazeer, Parmar, Uszkoreit, Jones,
  Gomez, Kaiser, and Polosukhin}]{vaswani2017attention}
Ashish Vaswani, Noam Shazeer, Niki Parmar, Jakob Uszkoreit, Llion Jones,
  Aidan~N Gomez, {\L}ukasz Kaiser, and Illia Polosukhin. 2017.
\newblock Attention is all you need.
\newblock In \emph{Advances in neural information processing systems}, pages
  5998--6008.

\bibitem[{Wu et~al.(2016)Wu, Schuster, Chen, Le, Norouzi, Macherey, Krikun,
  Cao, Gao, Macherey et~al.}]{wu2016google}
Yonghui Wu, Mike Schuster, Zhifeng Chen, Quoc~V Le, Mohammad Norouzi, Wolfgang
  Macherey, Maxim Krikun, Yuan Cao, Qin Gao, Klaus Macherey, et~al. 2016.
\newblock Google's neural machine translation system: Bridging the gap between
  human and machine translation.
\newblock \emph{arXiv preprint arXiv:1609.08144}.

\bibitem[{Yang et~al.(2019)Yang, Dai, Yang, Carbonell, Salakhutdinov, and
  Le}]{yang2019xlnet}
Zhilin Yang, Zihang Dai, Yiming Yang, Jaime Carbonell, Ruslan Salakhutdinov,
  and Quoc~V Le. 2019.
\newblock Xlnet: Generalized autoregressive pretraining for language
  understanding.
\newblock \emph{arXiv preprint arXiv:1906.08237}.

\end{thebibliography}
\bibliographystyle{acl_natbib}
\begin{appendix}

\section{Data Statistic}
Data statistics of the training/validation/testing data splits for six programming languages are given in Table \ref{table-codesearchnet-data-statistic}.
\begin{table}[ht]
\begin{center}
\begin{small}
\begin{sc}
\begin{tabular}{lccc}
\toprule
Code Search & Training & Dev & Testing \\
\midrule
Go&635,635&28,483&14,291\\
Java&908,886&30,655&26,909\\
JavaScript&247,773&16,505&6,483\\
PHP&1,047,406&52,029&28,391\\
Python&824,342&46,213&22,176\\
Ruby&97,580&4,417&2,279\\
\bottomrule
\end{tabular}
\end{sc}
\end{small}
\end{center}
\vskip -0.1in
\caption{ \label{table-codesearchnet-data-statistic} Data statistics about the CodeSearchNet Corpus for natural language code search.}
\end{table}
\section{Train Details}
\subsection{Pre-training}
We train CodeBERT on one NVIDIA DGX-2 machine using FP16. It combines 16 interconnected NVIDIA Tesla V100 with 32GB memory. We use the following set of hyper-parameters to train models: batchsize is 2,048 and learning rate is 5e-4. We use Adam to update the parameters and set the number of warmup steps as 10K. 
We set the max length as 512 and the max training step is 100K. Training 1,000 batches of data costs 600 minutes with MLM objective, 120 minutes with RTD objective.

\subsection{CodeSearch}
In the fine-turning step, we set the learning rate as 1e-5, the batch size as 64, the max sequence length as 200 and the max fine-tuning epoch as 8. 
As the same with pre-training, We use Adam to update the parameters. We choose the model performed best on the development set, and use that to evaluate on the test set.

\subsection{Code Summarization on Six Programming Languages}
We use Transformer with 6 layers, 768 dimensional hidden states and 12 attention heads as our decoder in all settings. We set the max length of input and inference as 256 and 64, respectively. We use the Adam optimizer to update model parameters. The learning rate and the batch size are 5e-5 and 64, respectively. We tune hyperparameters and perform early stopping on the development set.
\subsection{Code Summarization on C\#}
Since state-of-the-art methods use RNN as their decoder, we choose a 2-layer GRU with an attention mechanism as our decoder for a comparison.
We fine-tune  models using a grid search with the following set of hyper-parameters: batchsize is in \{32, 64\} and learning rate is in \{2e-5, 5e-5\}. We report the number when models achieve best performance on the development set. 

\section{Learning Curve of CodeSearch}\label{}
\begin{figure}[ht]
 \vskip 0.1in
\begin{center}
\centerline{\includegraphics[width=\columnwidth]{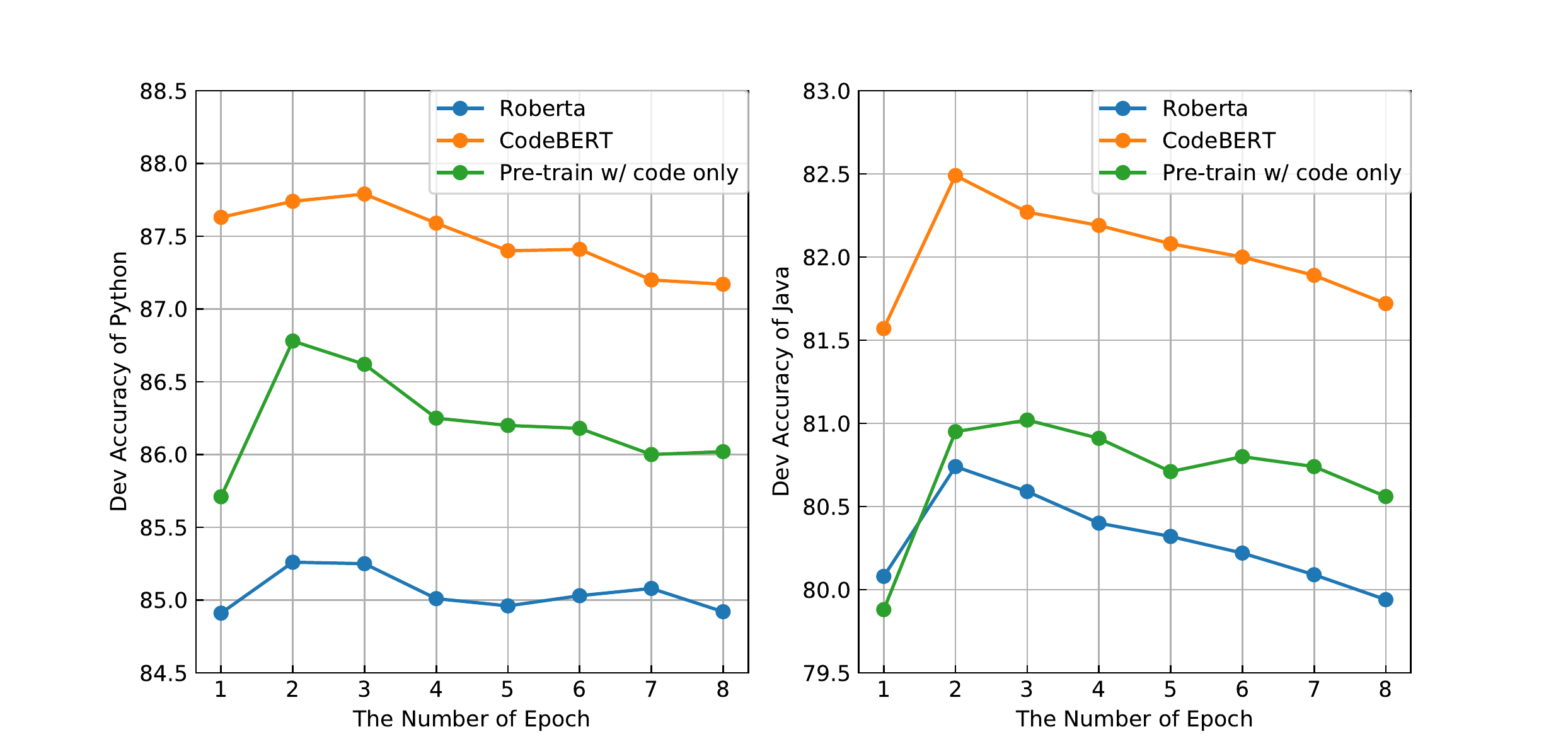}}
\caption{Learning curve of different pre-trained models in the fine-tuning step. We show results on Python and Java. }
\label{figure-curve}
\end{center}
\vskip -0.35in
\end{figure}

From Figure \ref{figure-curve}, we can see that CodeBERT performs better at the early stage, which reflects that CodeBERT provides good initialization for learning  downstream tasks.

\begin{figure*}
    \begin{center}
    \resizebox{0.94\linewidth}{!}{
    \includegraphics[scale=0.6]{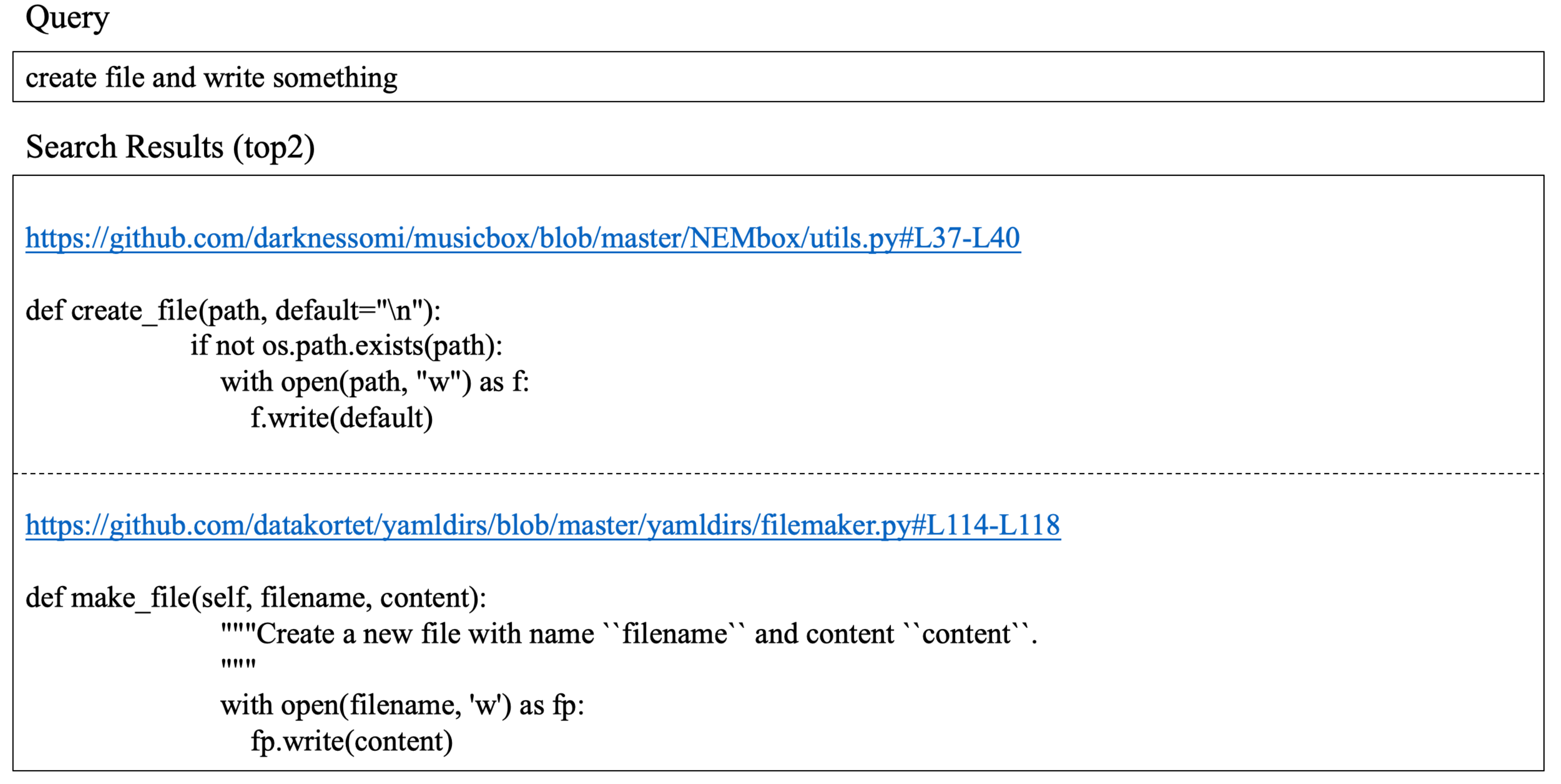}
    }
    \caption{Python CodeSearch example. The results are searched from 1,156,085 python code data. We only give the top2 results because space is limited.}
    \label{figure-codesearch}
    \end{center}

\end{figure*}

\begin{figure*}[ht]
    \begin{center}
    \resizebox{0.94\linewidth}{!}{
    \includegraphics[scale=0.6]{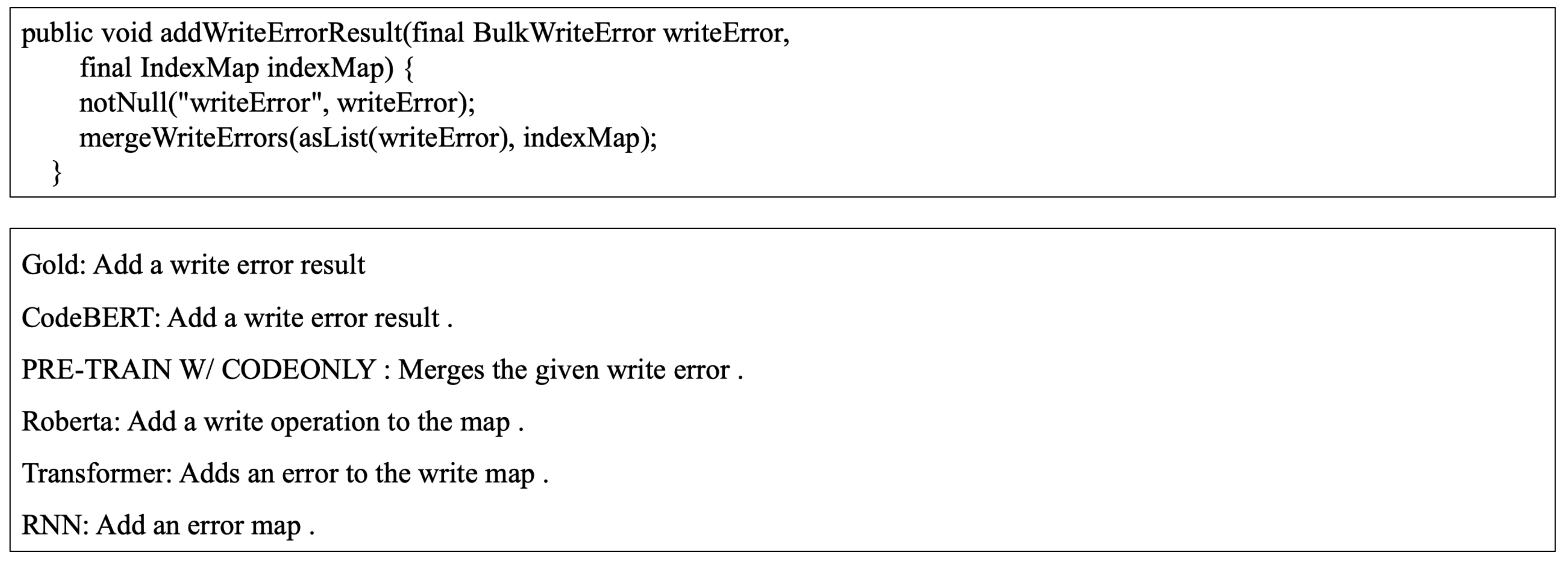}
    }
    \caption{Java code documentation generation output example. }
    \label{figure-java-generation}
    \end{center}
\end{figure*}

\section{Late Fusion}
In section { \S \ref{section:experiment-code-search}
}, we show that CodeBERT performs well in the setting where natural languages and codes have early interactions.
Here, we investigate whether CodeBERT is good at working as a unified encoder. 
We apply CodeBERT for natural language code search in a later fusion setting, where CodeBERT first encodes NL and PL separately, and then calculates the similarity by dot-product.
In this way, code search is equivalent to find the nearest codes in the shared vector space. This scenario also facilitates the use of CodeBERT in an online system, where the representations of codes are calculated in advance. In the runtime, a system only needs to compute the representation of NL and vector-based dot-product.

We fine-tune CodeBERT with the following objective, which maximizes the dot-product of the ground truth while minimizing the dot-product of distractors.
\begin{equation}
-\frac{1}{N}\sum_i \text{log}\bigg(\frac{\text{exp}\big(Enc(c_i)^\intercal Enc(w_i)\big)}{\sum_j\text{exp}\big(Enc(c_j)^\intercal Enc(w_i)\big)}\bigg)
\end{equation}
\begin{table}[ht]
	
	\vskip 0.1in
	\begin{center}
		\begin{small}
			\begin{sc}
				\begin{tabular}{lcc}
					\toprule
					model & Ruby & go\\
					\midrule
					RoBERTa& 0.0043 & 0.0030\\
 					Pre-Train w/ code only  & 0.1648 & 0.4179\\
					CodeBERT  & 0.6870 & 0.8372\\
 					\bottomrule
				\end{tabular}
			\end{sc}
 		\end{small}
 	\end{center}
 	\vskip -0.1in
 	\caption{\label{table-codesearchnet-dot-product} Results on natural language code search by late fusion. }
	
 \end{table} 

Results are given in Table \ref{table-codesearchnet-dot-product}. We just do this setting on two languages with a relatively small amount of data.

We can see that CodeBERT performs better than RoBERTa and the model pre-trained with codes only.  And late fusion performs comparable with the standard way. What's more, late fusion is more efficient and this setting could be used in an online system.

\section{Case Study}

\begin{figure*}[ht]
    \begin{center}
    \resizebox{0.94\linewidth}{!}{
    \includegraphics[scale=0.6]{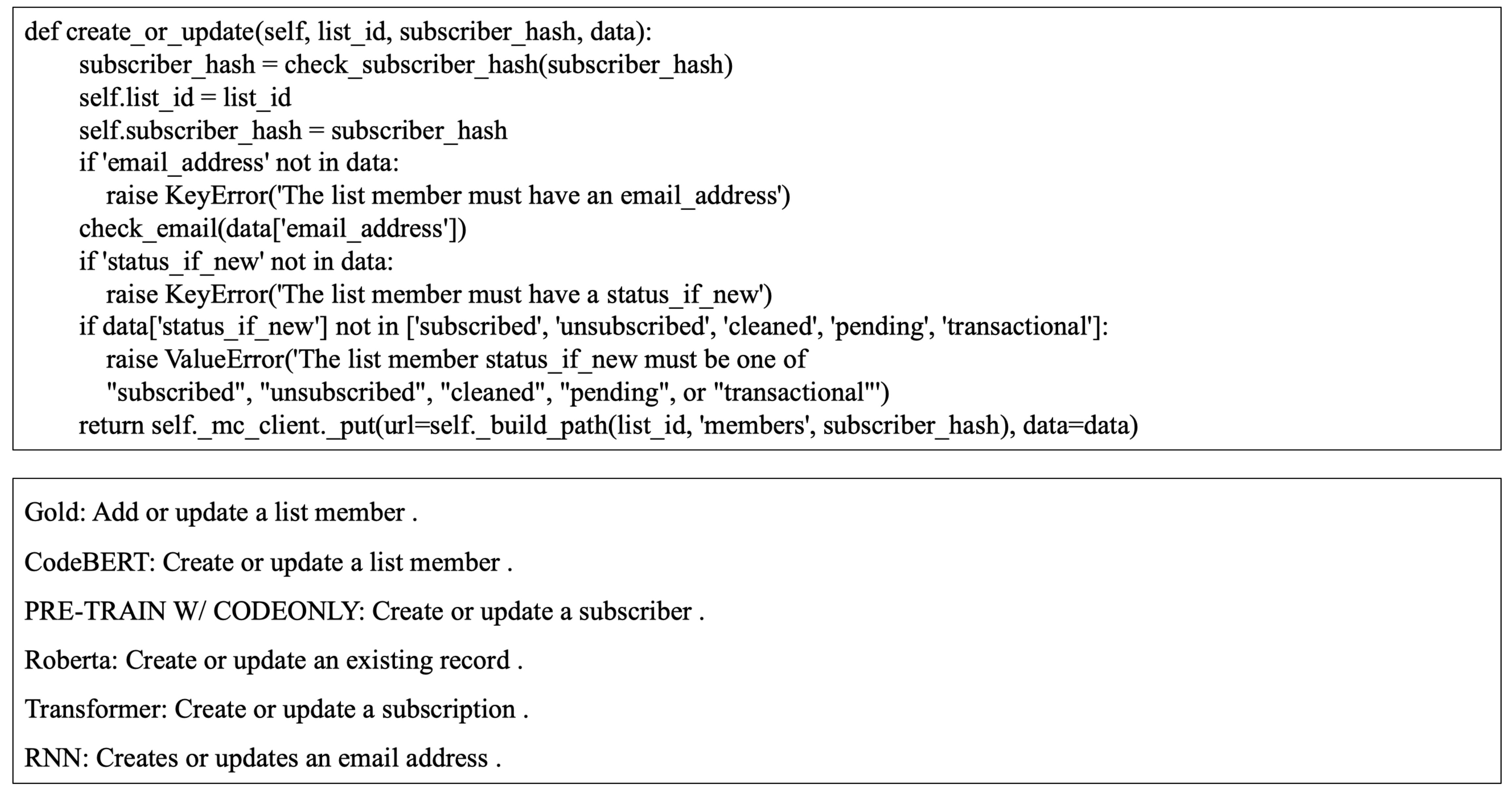}
    }
    \caption{Python code documentation generation output example. }
    \label{figure-python-generation}
    \end{center}
\end{figure*}

To qualitatively analyze the effectiveness of CodeBERT, we give some cases for code search and code documentation generation tasks. 

Considering the limited space, we only give the top2 results of the query for python programming language. As show in Figure \ref{figure-codesearch}, search results are very relevant with query. 

Figure \ref{figure-java-generation} and Figure \ref{figure-python-generation} show the outputs with different models for the code documentation generation task. As we can see, CodeBERT performs better than all baselines.

\end{appendix}

\end{document}